\documentclass[11pt]{article}
\usepackage{acl2014}
\usepackage{times}
\usepackage{url}
\usepackage{latexsym}

\usepackage{xspace}
\usepackage[export]{adjustbox} 
\usepackage{tikz}
\usepackage{pgfplots}
\usepackage{booktabs} 
\usepackage{amsmath}  
\usepackage{amsfonts} 
\usepackage{multirow}
\usepackage{caption} 

\newcommand{\addMod}{\textsc{Add}\xspace}
\newcommand{\addModplus}{\textsc{Add+}\xspace}
\newcommand{\flatMod}{\textsc{Bi}\xspace}
\newcommand{\flatModplus}{\textsc{Bi+}\xspace}
\newcommand{\docMod}{\textsc{Doc}\xspace}
\newcommand{\docModadd}{\textsc{Doc/Add}\xspace}
\newcommand{\docModflat}{\textsc{Doc/Bi}\xspace}

\newcommand{\CVM}{\textsc{CVM}\xspace}
\newcommand{\single}{\textit{single}\xspace}
\newcommand{\joint}{\textit{joint}\xspace}

\newcommand\kmh[1]{}

\title{Multilingual Models for Compositional Distributed Semantics}

\author{Karl Moritz Hermann \and Phil Blunsom\\
 Department of Computer Science\\
 University of Oxford\\
 Oxford, OX1 3QD, UK\\
 {\tt \{karl.moritz.hermann,phil.blunsom\}@cs.ox.ac.uk}}

\date{}

\begin{document}
\maketitle
\begin{abstract}
We present a novel technique for learning semantic representations, which
extends the distributional hypothesis to multilingual data and joint-space
embeddings. Our models leverage parallel data and learn to strongly align the
embeddings of semantically equivalent sentences, while maintaining sufficient
distance between those of dissimilar sentences. The models do not rely on
word alignments or any syntactic information and are successfully applied to a
number of diverse languages. We extend our approach to learn semantic
representations at the document level, too. We evaluate these models on two cross-lingual
document classification tasks, outperforming the prior state of the art.
Through qualitative analysis and the study of pivoting effects we demonstrate
that our representations are semantically plausible and can capture semantic
relationships across languages without parallel data.
\end{abstract}

\begin{figure}[t]
  \centering
  \captionsetup{font=small}
  \includegraphics[scale=0.55]{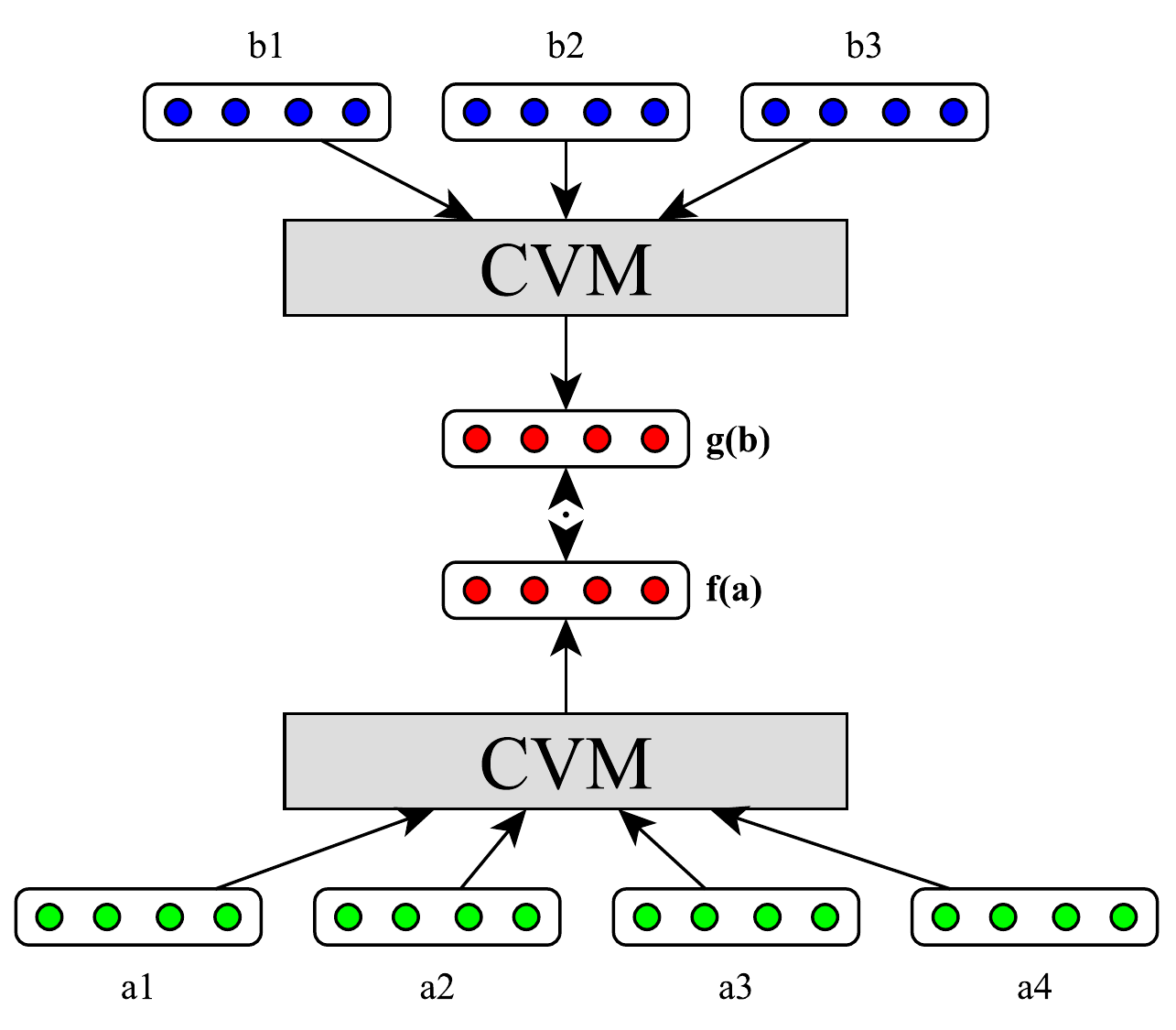}
  \caption{Model with parallel input sentences $a$ and $b$. The model
    minimises the distance between the sentence level encoding of the bitext.
    Any composition functions (\CVM) can be used to generate the
    compositional sentence level representations.
    }\label{fig:bilingual}
  \vspace{-1em}
\end{figure}
\section{Introduction}

Distributed representations of words provide the basis for many state-of-the-art
approaches to various problems in natural language processing today. Such word
embeddings are naturally richer representations than those of symbolic or
discrete models, and have been shown to be able to capture both syntactic and
semantic information. Successful applications of such models include language
modelling \cite{Bengio:2003}, paraphrase detection \cite{Erk:2008}, and dialogue
analysis \cite{Kalchbrenner:2013}.

Within a monolingual context, the distributional hypothesis \cite{Firth:1957}
forms the basis of most approaches for learning word representations. In this
work, we extend this hypothesis to multilingual data and joint-space embeddings.
We present a novel unsupervised technique for learning semantic
representations that leverages parallel corpora and employs semantic transfer
through compositional representations. Unlike most methods for learning word
representations, which are restricted to a single language, our approach learns to
represent meaning across languages in a shared multilingual semantic space.

We present experiments on two corpora.  First, we show that for cross-lingual
document classification on the Reuters RCV1/RCV2 corpora \cite{Lewis:2004}, we
outperform the prior state of the art \cite{Klementiev:2012}. Second, we also
present classification results on a massively multilingual corpus which we
derive from the TED corpus \cite{Cettolo:2012}. The results on this task, in
comparison with a number of strong baselines, further demonstrate the relevance
of our approach and the success of our method in learning multilingual semantic
representations over a wide range of languages.

\section{Overview}

Distributed representation learning describes the task of learning continuous
representations for discrete objects.  Here, we focus on learning semantic
representations and investigate how the use of multilingual data can improve
learning such representations at the word and higher level.  We
present a model that learns to represent each word in a lexicon by a continuous
vector in $\mathbb{R}^d$.  Such distributed representations allow a model to
share meaning between similar words, and have been used to capture semantic,
syntactic and morphological content \cite[\textit{inter
    alia}]{Collobert:2008,Turian:2010}.

We describe a multilingual objective function that uses a noise-contrastive
update between semantic representations of different languages to learn these
word embeddings. As part of this, we use a compositional vector model (\CVM,
  henceforth) to compute semantic representations of sentences and documents. A
\CVM learns semantic representations of larger syntactic units given the
semantic representations of their constituents \cite[\textit{inter
    alia}]{Clark:2007a,Mitchell:2008,Baroni:2010,Grefenstette:2011,Socher:2012,Hermann:2013:ACL}.

A key difference between our approach and those listed above is that we only
require sentence-aligned parallel data in our otherwise unsupervised learning
function. This removes a number of constraints that normally come with \CVM
models, such as the need for syntactic parse trees, word alignment or annotated
data as a training signal. At the same time, by using multiple \CVM{s} to transfer
information between languages, we enable our models to capture a broader
semantic context than would otherwise be possible.

The idea of extracting semantics from multilingual data stems from prior work in
the field of semantic grounding.  Language acquisition in humans is widely seen
as grounded in sensory-motor experience \cite{Bloom:2001,Roy:2003}.  Based on
this idea, there have been some attempts at using multi-modal data for learning
better vector representations of words (e.g. \newcite{Srivastava:2012}).  Such
methods, however, are not easily scalable across languages or to large amounts
of data for which no secondary or tertiary representation might exist.

Parallel data in multiple languages provides an alternative to such secondary
representations, as parallel texts share their semantics, and thus one language
can be used to ground the other.  Some work has exploited this idea for
transferring linguistic knowledge into low-resource languages or to learn
distributed representations at the word level \cite[\textit{inter
    alia}]{Klementiev:2012,Zou:2013,Lauly:2013}.  So far almost all of this
work has been focused on learning multilingual representations at the word
level.  As distributed representations of larger expressions have been shown to
be highly useful for a number of tasks, it seems to be a natural next step to
attempt to induce these, too, cross-lingually.

\section{Approach}

Most prior work on learning compositional semantic representations employs parse
trees on their training data to structure their composition functions
\cite[\textit{inter alia}]{Socher:2012,Hermann:2013:ACL}. Further, these
approaches typically depend on specific \textit{semantic} signals such as
sentiment- or topic-labels for their objective functions.  While these methods
have been shown to work in some cases, the need for parse trees and annotated
data limits such approaches to resource-fortunate languages. Our novel method
for learning compositional vectors removes these requirements, and as such can
more easily be applied to low-resource languages.

Specifically, we attempt to learn semantics from multilingual data. The idea is
that, given enough parallel data, a shared representation of two parallel
sentences would be forced to capture the common elements between these two
sentences. What parallel sentences share, of course, are their semantics.
Naturally, different languages express meaning in different ways. We utilise
this diversity to abstract further from mono-lingual surface realisations to
deeper semantic representations.  We exploit this semantic similarity across
languages by defining a bilingual (and trivially multilingual) energy as
follows.

Assume two functions ${f:X\rightarrow \mathbb{R}^d}$
and ${g:Y\rightarrow \mathbb{R}^d}$, which map sentences from languages $x$ and
$y$ onto distributed semantic representations in $\mathbb{R}^d$. Given a
parallel corpus $C$, we then define the energy of the model given two sentences
$(a,b) \in C$ as:
\begin{equation}
E_{bi}(a,b) = \left\| f(a) - g(b) \right\|^2\label{eqn:bi-error}
\end{equation}
We want to minimize $E_{bi}$ for all semantically equivalent sentences in the
corpus. In order to prevent the model from degenerating, we further introduce a
noise-constrastive large-margin update which ensures that the representations of
non-aligned sentences observe a certain margin from each other.
For every pair of parallel sentences $(a,b)$ we sample a number of additional
sentence pairs $(\cdot,n) \in C$, where $n$---with high probability---is not
semantically equivalent to $a$. We use these noise samples as follows:
\begin{equation}
E_{hl}(a,b,n) = \left[m + E_{bi}(a,b) - E_{bi}(a,n)\right]_{+}\nonumber
\end{equation}
where $[x]_{+} = max(x,0)$ denotes the standard hinge loss and $m$ is the margin.
This results in the following objective function:
\begin{equation}
J(\theta)=\sum_{(a,b) \in \mathcal{C}} \left( \sum_{i=1}^{k} E_{hl}(a,b,n_i) +
  \frac{\lambda}{2}\|\theta\|^2 \right)\label{eqn:objective}
\end{equation}
where $\theta$ is the set of all model variables.

\subsection{Two Composition Models}

The objective function in Equation \ref{eqn:objective} could be
coupled with any two given vector composition functions $f,g$ from the
literature. As we aim to apply our approach to a wide range of languages, we
focus on composition functions that do not require any syntactic information. We
evaluate the following two composition functions.

The first model, \addMod, represents a sentence by the sum of its word vectors.
This is a distributed bag-of-words approach as sentence ordering is not taken
into account by the model.

Second, the \flatMod model is designed to capture bigram information, using a
non-linearity over bigram pairs in its composition function:
\begin{equation}
  f(x) = \sum_{i=1}^{n} \text{tanh}\left(x_{i-1} + x_{i}\right)
\end{equation}
The use of a non-linearity enables the model to learn interesting interactions
between words in a document, which the bag-of-words approach of \addMod is not
capable of learning. We use the hyperbolic tangent as activation function.

\subsection{Document-level Semantics}\label{sec:docmod}

\begin{figure}[t]\centering
  \captionsetup{font=small}
  \includegraphics[scale=0.25]{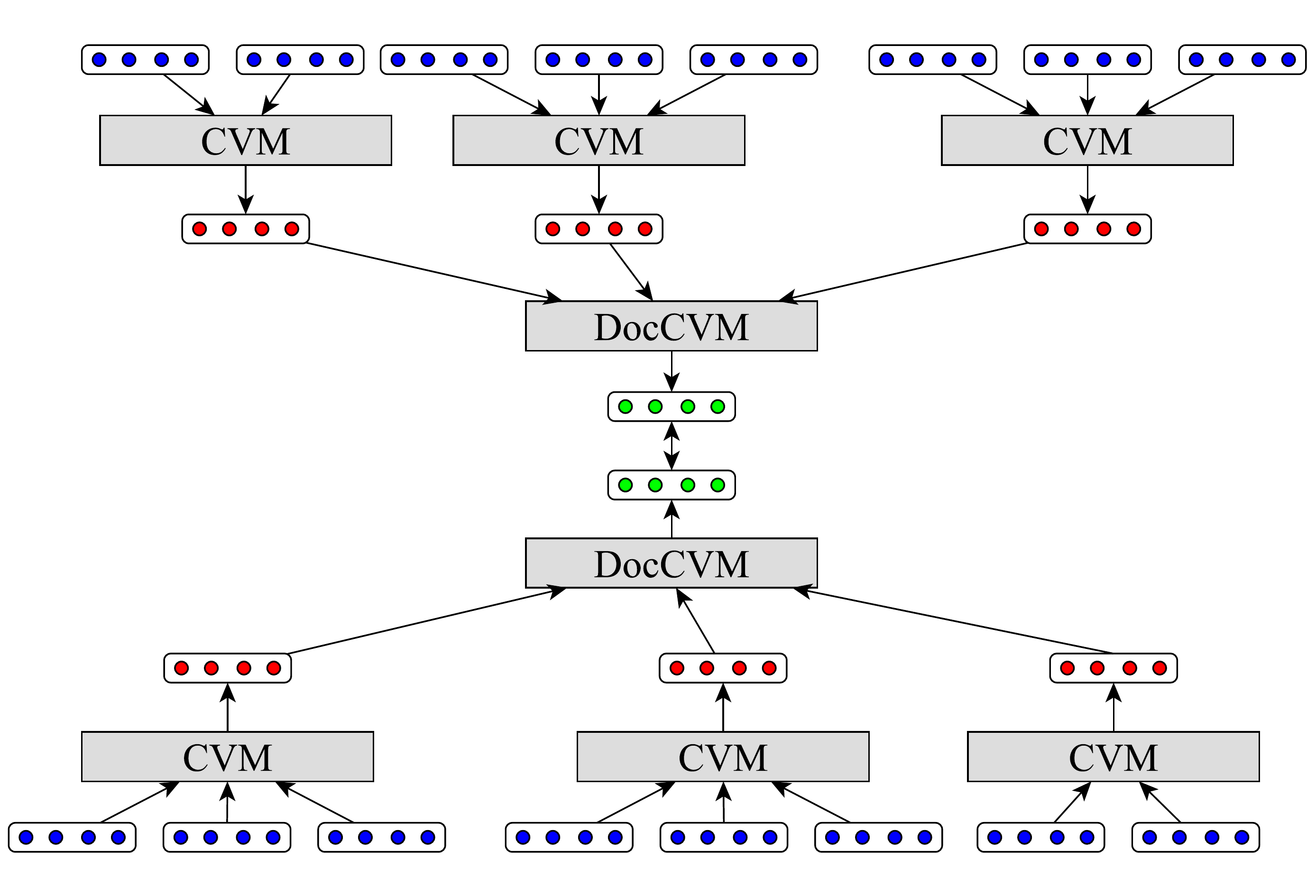}
  \caption{Description of a parallel document-level compositional vector model
    (\docMod). The model recursively computes semantic representations for each
    sentence of a document and then for the document itself, treating the sentence
    vectors as inputs for a second \CVM.
    }\label{fig:docmod}
\end{figure}

For a number of tasks, such as topic modelling, representations of objects beyond
the sentence level are required. While most approaches to compositional
distributed semantics end at the word level, our model extends to document-level
learning quite naturally, by recursively applying the composition and objective
function (Equation \ref{eqn:objective}) to compose sentences into documents.
This is achieved by first computing semantic representations for each sentence
in a document.  Next, these representations are used as inputs in a higher-level
\CVM, computing a semantic representation of a document (Figure
  \ref{fig:docmod}).

This recursive approach integrates document-level representations into the
learning process. We can thus use corpora of parallel documents---regardless of
whether they are sentence aligned or not---to propagate a semantic signal back
to the individual words.
If sentence alignment is available, of course, the document-signal can simply be
combined with the sentence-signal, as we did with the experiments described in
\S\ref{sec:ted-cldc}.

This concept of learning compositional representations for documents contrasts
with prior work \cite[\textit{inter alia}]{Socher:2011,Klementiev:2012} who rely
on summing or averaging sentence-vectors if representations beyond the
sentence-level are required for a particular task.

We evaluate the models presented in this paper both with and without the
document-level signal. We refer to the individual models used as \addMod and
\flatMod if used without, and as \docModadd and \docModflat is used with the
additional document composition function and error signal.

\section{Corpora}\label{sec:corpus}

We use two corpora for learning semantic representations and performing the
experiments described in this paper.

The Europarl corpus v7\footnote{\url{http://www.statmt.org/europarl/}}
\cite{Koehn:2005} was used during initial development and testing of our
approach, as well as to learn the representations used for the Cross-Lingual
Document Classification task described in \S\ref{sec:rcv-cldc}.  We considered
the English-German and English-French language pairs from this corpus.  From
each pair the final 100,000 sentences were reserved for development.

Second, we developed a massively multilingual corpus based on the TED
corpus\footnote{\url{https://wit3.fbk.eu/}} for IWSLT 2013 \cite{Cettolo:2012}.
This corpus contains English transcriptions and multilingual, sentence-aligned
translations of talks from the TED conference.  While the corpus is aimed at
machine translation tasks, we use the keywords associated with each talk to
build a subsidiary corpus for multilingual document classification as
follows.\footnote{\url{http://www.clg.ox.ac.uk/tedcldc/}}

The development sections provided with the IWSLT 2013 corpus were again reserved
for development.  We removed approximately 10 percent of the training data in
each language to create a test corpus (all talks with $id\geq1{,}400$).  The new
training corpus consists of a total of 12,078 parallel documents distributed
across 12 language pairs\footnote{English to Arabic, German, French, Spanish,
  Italian, Dutch, Polish, Brazilian Portuguese, Romanian, Russian and Turkish.
  Chinese, Farsi and Slowenian were removed due to the small size of those
  datasets.}. In total, this amounts to 1,678,219 non-English sentences (the
  number of unique English sentences is smaller as many documents are translated
  into multiple languages and thus appear repeatedly in the corpus).  Each
document (talk) contains one or several keywords.  We used the 15 most frequent
keywords for the topic classification experiments described in section
\S\ref{sec:ted-cldc}.

Both corpora were pre-processed using the set of tools provided by
cdec\footnote{\url{http://cdec-decoder.org/}} for tokenizing and
lowercasing the data.  Further, all empty sentences and their translations were
removed from the corpus.

\section{Experiments}

We report results on two experiments. First, we replicate the cross-lingual
document classification task of \newcite{Klementiev:2012}, learning distributed
representations on the Europarl corpus and evaluating on documents from the
Reuters RCV1/RCV2 corpora.  Subsequently, we design a multi-label classification
task using the TED corpus, both for training and evaluating.  The use of a wider
range of languages in the second experiments allows
us to better evaluate our models' capabilities in
learning a shared multilingual semantic representation.  We also investigate the
learned embeddings from a qualitative perspective in \S\ref{sec:qualitative}.

\subsection{Learning}

All model weights were randomly initialised using a Gaussian distribution
($\mu{=}0,\sigma^2{=}0.1$). We used the available development data to set our
model parameters. For each positive sample we used a number of noise samples ($k
  \in \{1,10,50\}$), randomly drawn from the corpus at each training epoch.  All
our embeddings have dimensionality $d{=}128$, with the margin set to
$m{=}d$.\footnote{On the RCV task we also report results for $d{=}40$ which
  matches the dimensionality of \newcite{Klementiev:2012}.} Further, we use L2
regularization with $\lambda{=}1$ and step-size in $\{0.01, 0.05\}$. We use 100
iterations for the RCV task, 500 for the TED single and 5 for the joint
corpora. We use the adaptive gradient method, AdaGrad \cite{Duchi:2011}, for
updating the weights of our models, in a mini-batch setting ($b \in \{10,50\}$).
All settings, our model implementation and scripts to replicate our
experiments are available at \mbox{\url{http://www.karlmoritz.com/}}.

\subsection{RCV1/RCV2 Document Classification}\label{sec:rcv-cldc}

We evaluate our models on the cross-lingual document classification (CLDC,
  henceforth) task first described in \newcite{Klementiev:2012}.  This task
involves learning language independent embeddings which are then used for
document classification across the English-German language pair.  For this, CLDC
employs a particular kind of supervision, namely using supervised training data
in one language and evaluating without further supervision in another.  Thus,
CLDC can be used to establish whether our learned representations are
semantically useful across multiple languages.

We follow the experimental setup described in \newcite{Klementiev:2012}, with
the exception that we learn our embeddings using solely the Europarl data and
use the Reuters corpora only during for classifier training and testing.  Each
document in the classification task is represented by the average of the
$d$-dimensional representations of all its sentences.  We train the multiclass
classifier using an averaged perceptron \cite{Collins:2002} with the same
settings as in \newcite{Klementiev:2012}.

We present results from four models.  The \addMod model is trained on 500k
sentence pairs of the English-German parallel section of the Europarl corpus.
The \addModplus model uses an additional 500k parallel sentences from the
English-French corpus, resulting in one million English sentences, each paired
up with either a German or a French sentence, with \flatMod and \flatModplus
trained accordingly.
The motivation behind \addModplus and \flatModplus is to investigate whether we
can learn better embeddings by introducing additional data from other languages.
A similar idea exists in machine translation where English
is frequently used to pivot between other languages \cite{Cohn:2007}.

\begin{table}[t]\centering
  \captionsetup{font=small}
\begin{tabular}{@{}lrr@{}}\toprule
  \multicolumn{1}{@{}l}{Model} & en $\rightarrow$ de & de $\rightarrow$ en \\ \midrule
  \multicolumn{1}{@{}l}{Majority Class} & 46.8 & 46.8 \\
  \multicolumn{1}{@{}l}{Glossed} & 65.1 & 68.6 \\
  \multicolumn{1}{@{}l}{MT} & 68.1 & 67.4 \\
  \multicolumn{1}{@{}l}{I-Matrix} & 77.6 & 71.1 \\
  \midrule
  $dim = 40$ \\
  \addMod & 83.7 & 71.4 \\
  \addModplus & 86.2 & 76.9 \\
  \flatMod & 83.4 & 69.2 \\
  \flatModplus & 86.9 & 74.3 \\
  \midrule
  $dim = 128$ \\
  \addMod & 86.4 & 74.7 \\
  \addModplus & 87.7 & 77.5 \\
  \flatMod & 86.1 & 79.0 \\
  \flatModplus & \textbf{88.1} & \textbf{79.2} \\
  \bottomrule
\end{tabular}
\caption{Classification accuracy for training on English and German with 1000
  labeled examples on the RCV corpus. Cross-lingual compositional representations (\addMod,
    \flatMod and their multilingual extensions), I-Matrix \cite{Klementiev:2012}
  translated (MT) and glossed (Glossed) word baselines, and the majority class
  baseline. The baseline results are from \newcite{Klementiev:2012}.}
\label{tab:results1k}
\end{table}

\pgfplotsset{every axis plot/.append style={line width=1pt}}

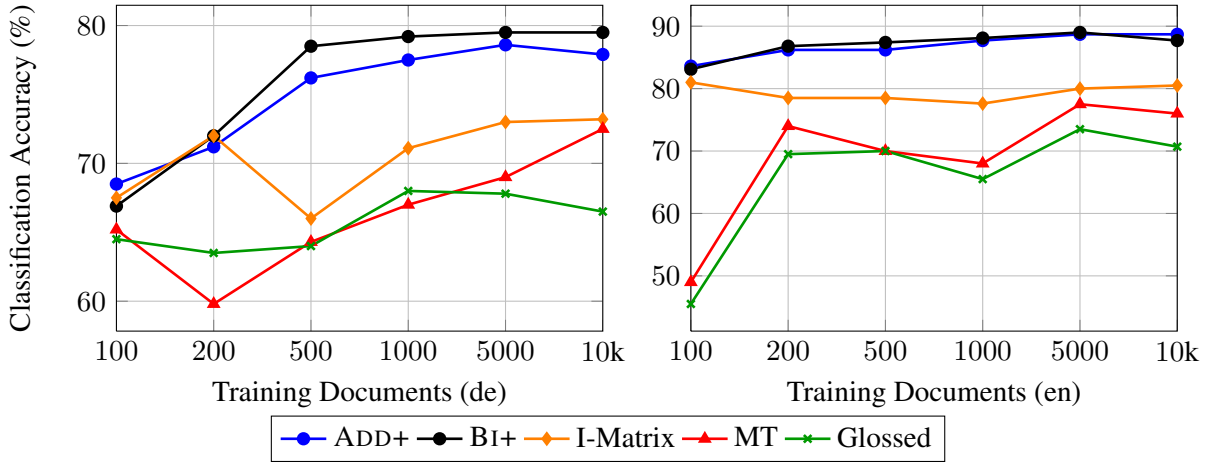
\begin{figure*}[t]
  \captionsetup{font=small}
\begin{tikzpicture}
	\begin{axis}[
		grid=major,
		ytick={40,50,...,80},
		xlabel=Training Documents (de),
    		xtick={1,2,3,4,5,6},
	    xticklabels={$100$,$200$,$500$,$1000$,$5000$,$10$k},
		ylabel=Classification Accuracy (\%),
		legend columns=-1,
		legend entries={\addModplus, \flatModplus, I-Matrix, MT, Glossed},
		legend to name=sharedlegend,
		scale only axis,
		xmin=1,xmax=6,
		width=0.4\textwidth,
		height=0.27\textwidth,
		]
	\addplot[color=blue,mark=*] coordinates {
	(1, 68.5)
	(2, 71.2)
	(3, 76.2)
	(4, 77.5)
	(5, 78.6)
	(6, 77.9)
	};
	\addplot[color=black,mark=*] coordinates {
	(1, 66.9)
	(2, 72.0)
	(3, 78.5)
	(4, 79.2)
	(5, 79.5)
	(6, 79.5)
	};
	\addplot[color=orange,mark=diamond*] coordinates {
	(1, 67.5)
	(2, 72)
	(3, 66)
	(4, 71.1)
	(5, 73)
	(6, 73.2)
	};
	\addplot[color=red,mark=triangle*] coordinates {
	(1, 65.2)
	(2, 59.8)
	(3, 64.3)
	(4, 67)
	(5, 69)
	(6, 72.5)
	};
	\addplot[color=green!60!black,mark=x] coordinates {
	(1, 64.5)
	(2, 63.5)
	(3, 64)
	(4, 68)
	(5, 67.8)
	(6, 66.5)
	};
	\end{axis}
\end{tikzpicture}
\begin{tikzpicture}
	\begin{axis}[
		grid=major,
		xlabel=Training Documents (en),
    		xtick={1,2,3,4,5,6},
	    xticklabels={$100$,$200$,$500$,$1000$,$5000$,$10$k},
		scale only axis,
		xmin=1,xmax=6,
    ytick={40,50,60,70,80,90},
		width=0.4\textwidth,
		height=0.27\textwidth,
		]
	\addplot[color=blue,mark=*] coordinates {
	(1, 83.6)
	(2, 86.2)
	(3, 86.2)
	(4, 87.7)
	(5, 88.7)
	(6, 88.7)
	};
	\addplot[color=black,mark=*] coordinates {
	(1, 83.1)
	(2, 86.8)
	(3, 87.4)
	(4, 88.1)
	(5, 89.0)
	(6, 87.7)
	};
	\addplot[color=orange,mark=diamond*] coordinates {
	(1, 81)
	(2, 78.5)
	(3, 78.5)
	(4, 77.6)
	(5, 80)
	(6, 80.5)
	};
	\addplot[color=red,mark=triangle*] coordinates {
	(1, 49)
	(2, 74)
	(3, 70)
	(4, 68)
	(5, 77.5)
	(6, 76)
	};
	\addplot[color=green!60!black,mark=x] coordinates {
	(1, 45.5)
	(2, 69.5)
	(3, 70)
	(4, 65.5)
	(5, 73.5)
	(6, 70.7)
	};
	\end{axis}
\end{tikzpicture}
\vspace{-0.5em}
\begin{center}
\ref{sharedlegend}
\end{center}
\vspace{-0.5em}
\caption{Classification accuracy for a number of models (see Table
    \ref{tab:results1k} for model descriptions). The left chart shows results
  for these models when trained on German data and evaluated on English data,
  the right chart vice versa.}\label{fig:cldccharts}
\end{figure*}

The actual CLDC experiments are performed by training on English and testing on
German documents and vice versa. Following prior work, we use varying sizes
between 100 and 10,000 documents when training the multiclass classifier. The
results of this task across training sizes are in Figure \ref{fig:cldccharts}.
Table \ref{tab:results1k} shows the results for training on 1,000 documents
compared with the results published in \newcite{Klementiev:2012}. Our models
outperform the prior state of the art, with the \flatMod models performing
slightly better than the \addMod models. As the relative results indicate, the
addition of a second language improves model performance. It it interesting to
note that results improve in both directions of the task, even though no
additional German data was used for the `+` models.

\subsection{TED Corpus Experiments}\label{sec:ted-cldc}

\begin{table*}\centering\small
\begin{tabular}{@{}lr@{\hspace{0.5em}}r@{\hspace{0.5em}}r@{\hspace{0.5em}}r@{\hspace{0.5em}}r@{\hspace{0.5em}}r@{\hspace{0.5em}}r@{\hspace{0.5em}}r@{\hspace{0.5em}}r@{\hspace{0.5em}}r@{\hspace{0.5em}}r@{}}\toprule
Setting & \multicolumn{11}{c}{Languages}\\ \cmidrule{2-12}
& Arabic & German & Spanish & French & Italian & Dutch & Polish & Pt-Br & Roman. & Russian & Turkish  \\
\midrule
\multicolumn{12}{@{}l}{$en\rightarrow \text{L2}$} \\
MT System &
\textbf{0.429} & \textbf{0.465} & \textbf{0.518} & \textbf{0.526} & \textbf{0.514} & \textbf{0.505} & \textbf{0.445} & \textbf{0.470} & \textbf{0.493} & 0.432 & 0.409 \\
\addMod \single &
0.328 & 0.343 & 0.401 & 0.275 & 0.282 & 0.317 & 0.141 & 0.227 & 0.282 & 0.338 & 0.241 \\
\flatMod \single &
0.375 & 0.360 & 0.379 & 0.431 & 0.465 & 0.421 & \underline{0.435} & 0.329 &
0.426 & 0.423 & \textbf{\underline{0.481}} \\
\docModadd \single &
\underline{0.410} & 0.424 & 0.383 & \underline{0.476} & \underline{0.485} & 0.264 & 0.402 & 0.354 & 0.418 & 0.448 & 0.452 \\
\docModflat \single &
0.389 & \underline{0.428} & 0.416 & 0.445 & 0.473 & 0.219 & 0.403 & 0.400 & \underline{0.467} & 0.421 & 0.457 \\
\docModadd \joint &
0.392 & 0.405 & 0.443 & 0.447 & 0.475 & \underline{0.453} & 0.394 &
\underline{0.409} & 0.446 & \textbf{\underline{0.476}} & 0.417 \\
\docModflat \joint &
0.372 & 0.369 & \underline{0.451} & 0.429 & 0.404 & 0.433 & 0.417 & 0.399 & 0.453 & 0.439 & 0.418 \\
\midrule
\multicolumn{12}{@{}l}{$\text{L2}\rightarrow en$} \\
MT System &
0.448 & 0.469 & \textbf{0.486} & 0.358 & \textbf{0.481} & 0.463 & \textbf{0.460} & 0.374 & \textbf{0.486} & 0.404 & 0.441 \\
\addMod \single &
0.380 & 0.337 & \underline{0.446} & 0.293 & 0.357 & 0.295 & 0.327 & 0.235 & 0.293 & 0.355 & 0.375 \\
\flatMod \single &
0.354 & 0.411 & 0.344 & 0.426 & 0.439 & 0.428 & \underline{0.443} &
0.357 & 0.426 & 0.442 & 0.403 \\
\docModadd \single &
\textbf{\underline{0.452}} & \textbf{\underline{0.476}} & 0.422 & 0.464 & \underline{0.461} & 0.251 & 0.400 & 0.338 & 0.407 & \underline{\textbf{0.471}} & 0.435 \\
\docModflat \single &
0.406 & 0.442 & 0.365 & \underline{\textbf{0.479}} & 0.460 & 0.235 & 0.393 & 0.380 & 0.426 & 0.467 & \underline{\textbf{0.477}} \\
\docModadd \joint &
0.396 & 0.388 & 0.399 & 0.415 & \underline{0.461} & \underline{\textbf{0.478}} &
0.352 & \textbf{\underline{0.399}} & 0.412 & 0.343 & 0.343 \\
\docModflat \joint &
0.343 & 0.375 & 0.369 & 0.419 & 0.398 & 0.438 & 0.353 & 0.391 &
\underline{0.430} & 0.375 & 0.388 \\
\bottomrule
\end{tabular}
\caption{F1-scores for the TED document classification task for individual
  languages. Results are reported for both directions (training on English,
    evaluating on L2 and vice versa). Bold indicates best result, underline best
  result amongst the vector-based systems.}\label{tab:exp-beta}
\end{table*}

\begin{table*}\centering\small
\begin{tabular}{@{}lr@{\hspace{0.4em}}r@{\hspace{0.4em}}r@{\hspace{0.4em}}r@{\hspace{0.4em}}r@{\hspace{0.4em}}r@{\hspace{0.4em}}r@{\hspace{0.4em}}r@{\hspace{0.4em}}r@{\hspace{0.4em}}r@{\hspace{0.4em}}r@{}}\toprule
\multirow{2}{*}{\begin{tabular}[l]{@{}l@{}}Training\\Language\end{tabular}} &
  \multicolumn{11}{c}{Test Language}\\ \cmidrule{2-12}
& Arabic & German & Spanish & French & Italian & Dutch & Polish & ~~Pt-Br & ~Rom'n & Russian & Turkish  \\
\midrule
Arabic     &       & 0.378 & 0.436 & 0.432 & 0.444 & 0.438 & 0.389 & 0.425 & 0.420 & 0.446 & 0.397 \\
German     & 0.368 &       & 0.474 & 0.460 & 0.464 & 0.440 & 0.375 & 0.417 & 0.447 & 0.458 & 0.443 \\
Spanish    & 0.353 & 0.355 &       & 0.420 & 0.439 & 0.435 & 0.415 & 0.390 & 0.424 & 0.427 & 0.382 \\
French     & 0.383 & 0.366 & 0.487 &       & 0.474 & 0.429 & 0.403 & 0.418 & 0.458 & 0.415 & 0.398 \\
Italian    & 0.398 & 0.405 & 0.461 & 0.466 &       & 0.393 & 0.339 & 0.347 & 0.376 & 0.382 & 0.352 \\
Dutch      & 0.377 & 0.354 & 0.463 & 0.464 & 0.460 &       & 0.405 & 0.386 & 0.415 & 0.407 & 0.395 \\
Polish     & 0.359 & 0.386 & 0.449 & 0.444 & 0.430 & 0.441 &       & 0.401 & 0.434 & 0.398 & 0.408 \\
Portuguese & 0.391 & 0.392 & 0.476 & 0.447 & 0.486 & 0.458 & 0.403 &       & 0.457 & 0.431 & 0.431 \\
Romanian   & 0.416 & 0.320 & 0.473 & 0.476 & 0.460 & 0.434 & 0.416 & 0.433 &       & 0.444 & 0.402 \\
Russian    & 0.372 & 0.352 & 0.492 & 0.427 & 0.438 & 0.452 & 0.430 & 0.419 & 0.441 &       & 0.447 \\
Turkish    & 0.376 & 0.352 & 0.479 & 0.433 & 0.427 & 0.423 & 0.439 & 0.367 & 0.434 & 0.411 &       \\
\bottomrule
\end{tabular}
\caption{F1-scores for TED corpus document classification results when training
  and testing on two languages that do not share any parallel data. We train a
  \docModadd model on all $en$-L2 language pairs together, and then use the
  resulting embeddings to train document classifiers in each language. These
  classifiers are subsequently used to classify data from all other
  languages.}\label{tab:exp-gamma}
\end{table*}

\begin{table*}\centering\small
\begin{tabular}{@{}lr@{\hspace{0.5em}}r@{\hspace{0.5em}}r@{\hspace{0.5em}}r@{\hspace{0.5em}}r@{\hspace{0.5em}}r@{\hspace{0.5em}}r@{\hspace{0.5em}}r@{\hspace{0.5em}}r@{\hspace{0.5em}}r@{\hspace{0.5em}}r@{\hspace{0.5em}}r@{}}\toprule
Setting & \multicolumn{12}{c}{Languages}\\ \cmidrule{2-13}
& English & Arabic & German & Spanish & French & Italian & Dutch & Polish & Pt-Br & Roman. & Russian & Turkish  \\
\midrule
Raw Data NB
& 0.481 & 0.469 & 0.471 & 0.526 & 0.532 & 0.524 & 0.522 & 0.415 & 0.465 & 0.509 & 0.465 & 0.513 \\
\midrule
Senna	& 0.400 & & & & & & & & & & & \\ 
Polyglot
& 0.382 & 0.416 & 0.270 & 0.418 & 0.361 & 0.332 & 0.228 & 0.323 & 0.194 & 0.300 & 0.402 & 0.295 \\
\midrule
\single Setting \\
\docModadd &
0.462 & 0.422 & 0.429 & 0.394 & 0.481 & 0.458 & 0.252 & 0.385 & 0.363 & 0.431 & 0.471 & 0.435 \\
\docModflat &
0.474 & 0.432 & 0.362 & 0.336 & 0.444 & 0.469 & 0.197 & 0.414 & 0.395 & 0.445 & 0.436 & 0.428 \\
\joint Setting \\
\docModadd &
0.475 & 0.371 & 0.386 & 0.472 & 0.451 & 0.398 & 0.439 & 0.304 & 0.394 & 0.453 & 0.402 & 0.441 \\
\docModflat &
0.378 & 0.329 & 0.358 & 0.472 & 0.454 & 0.399 & 0.409 & 0.340 & 0.431 & 0.379 & 0.395 & 0.435 \\
\bottomrule
\end{tabular}
\caption{F1-scores on the TED corpus document classification task when training
  and evaluating on the same language. Baseline embeddings are Senna
  \cite{Collobert:2011} and Polyglot \cite{Al-Rfou:2013}.}\label{tab:exp-delta}
\end{table*}

Here we describe our experiments on the TED corpus, which enables us to scale up
to multilingual learning. Consisting of a large number of relatively short and
parallel documents, this corpus allows us to evaluate the performance of
the \docMod model described in \S\ref{sec:docmod}.

We use the training data of the corpus to learn distributed representations
across 12 languages. Training is performed in two settings. In the \single
mode, vectors are learnt from a single language pair (en-X), while in the \joint
mode vector-learning is performed on all parallel sub-corpora simultaneously.
This setting causes words from all languages to be embedded in a single semantic
space.

First, we evaluate the effect of the document-level error signal (\docMod,
  described in \S\ref{sec:docmod}), as well as whether our multilingual learning
method can extend to a larger variety of languages. We train \docMod models,
using both \addMod and \flatMod as \CVM (\docModadd, \docModflat), both in the
\single and \joint mode. For comparison, we also train \addMod and \docMod
models without the document-level error signal.  The resulting document-level
representations are used to train classifiers (system and settings as in
  \S\ref{sec:rcv-cldc}) for each language, which are then evaluated in the
paired language. In the English case we train twelve individual classifiers,
each using the training data of a single language pair only.  As described in
\S\ref{sec:corpus}, we use 15 keywords for the classification task.  Due to
space limitations, we report cumulative results in the form of F1-scores
throughout this paper.

\vspace{0.05in}
\noindent\textbf{MT System}\hspace{0.1in}
We develop a machine translation baseline as follows.  We train a
machine translation tool on the parallel training data, using the development
data of each language pair to optimize the translation system.  We use the cdec
decoder \cite{Dyer:2010} with default settings for this purpose.  With this
system we translate the test data, and then use a Na\"{i}ve Bayes
classifier\footnote{We use the implementation in Mallet \cite{McCallum:2002}}
for the actual experiments.  To exemplify, this means the $de{\to}ar$ result is
produced by training a translation system from Arabic to German. The Arabic
test set is translated into German. A classifier is then trained on the German
training data and evaluated on the translated Arabic. While we developed this
system as a baseline, it must be noted that the classifier of this system has
access to significantly more information (all words in the document) as opposed
to our models (one embedding per document), and we do not expect to necessarily
beat this system.

The results of this experiment are in Table \ref{tab:exp-beta}.  When comparing
the results between the \addMod model and the models trained using the
document-level error signal, the benefit of this additional signal becomes
clear. The \joint training mode leads to a relative improvement when training
on English data and evaluating in a second language.  This suggests that the
\joint mode improves the quality of the English embeddings more than it affects
the L2-embeddings. More surprising, perhaps, is the relative performance
between the \addMod and \flatMod composition functions, especially when compared
to the results in \S\ref{sec:rcv-cldc}, where the \flatMod models relatively
consistently performed better. We suspect that the better performance of the
additive composition function on this task is related to the smaller amount of
training data available which could cause sparsity issues for the bigram model.

As expected, the MT system slightly outperforms our models on most language
pairs. However, the overall performance of the models is comparable to that of
the MT system. Considering the relative amount of information available during
the classifier training phase, this indicates that our learned representations
are semantically useful, capturing almost the same amount of information as
available to the Na\"{i}ve Bayes classifier.

We next investigate linguistic transfer across languages.  We re-use
the embeddings learned with the \docModadd \joint model from the previous
experiment for this purpose, and train classifiers on all non-English languages
using those embeddings. Subsequently, we evaluate their performance in
classifying documents in the remaining languages.  Results for this task are in
Table \ref{tab:exp-gamma}. While the results across language-pairs might not be
very insightful, the overall good performance compared with the results in Table
\ref{tab:exp-beta} implies that we learnt semantically meaningful vectors and in
fact a joint embedding space across thirteen languages.

In a third evaluation (Table \ref{tab:exp-delta}), we apply the embeddings
learnt with out models to a monolingual classification task, enabling us to
compare with prior work on distributed representation learning.  In this
experiment a classifier is trained in one language and then evaluated in the
same. We again use a Na\"{i}ve Bayes classifier on the raw data to establish a
reasonable upper bound.

We compare our embeddings with the SENNA embeddings, which achieve state of the
art performance on a number of tasks \cite{Collobert:2011}. Additionally, we
use the Polyglot embeddings of \newcite{Al-Rfou:2013}, who published word
embeddings across 100 languages, including all languages considered in this
paper. We represent each document by the mean of its word vectors and then
apply the same classifier training and testing regime as with our models. Even
though both of these sets of embeddings were trained on much larger datasets
than ours, our models outperform these baselines on all languages---even
outperforming the Na\"{i}ve Bayes system on on several languages. While this
may partly be attributed to the fact that our vectors were learned on in-domain
data, this is still a very positive outcome.

\subsection{Linguistic Analysis}\label{sec:qualitative}

\begin{figure}[t]\centering
  \captionsetup{font=small}
\includegraphics[scale=0.35, frame]{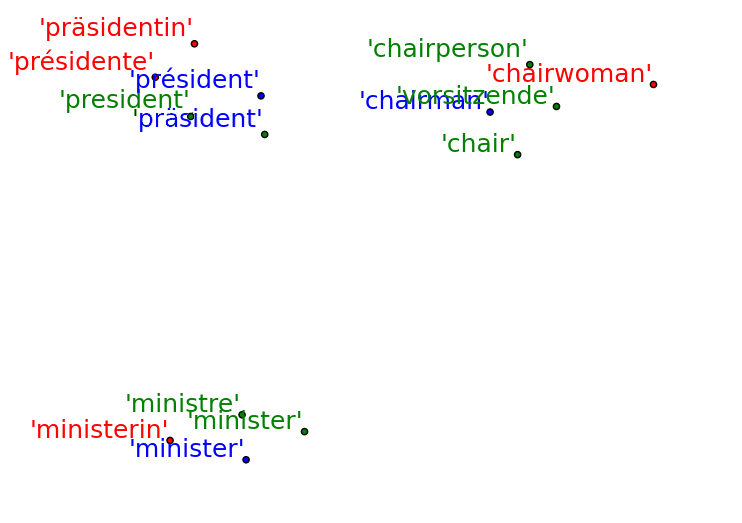}
\caption{t-SNE projections for a number of English, French and German words as
  represented by the \flatModplus model. Even though the model did not use any
  parallel French-German data during training, it learns semantic similarity
  between these two languages using English as a pivot, and semantically
  clusters words across all languages.}\label{fig:words}
\end{figure}

\begin{figure}[t]\centering
  \captionsetup{font=small}
\includegraphics[scale=0.35, frame]{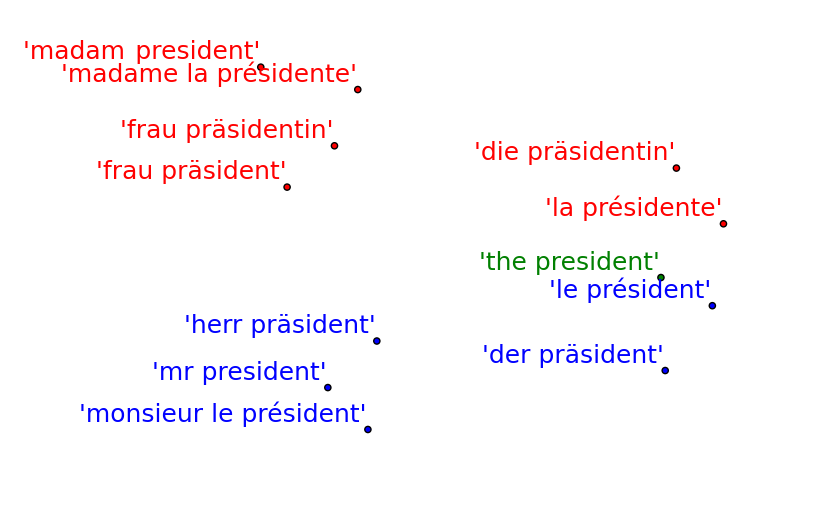}
\caption{t-SNE projections for a number of short phrases in three languages as
  represented by the \flatModplus model. The projection demonstrates linguistic
  transfer through a pivot by. It separates phrases by gender (red for female,
    blue for male, and green for neutral) and aligns matching phrases across
  languages.}\label{fig:gender}
\end{figure}

While the classification experiments focused on establishing the semantic
content of the sentence level representations, we also want to briefly
investigate the induced word embeddings.  We use the \flatModplus model trained
on the Europarl corpus for this purpose.  Figure \ref{fig:words} shows the t-SNE
projections for a number of English, French and German words.  Even though the
model did not use any parallel French-German data during training, it still
managed to learn semantic word-word similarity across these two languages.

Going one step further, Figure \ref{fig:gender} shows t-SNE projections for a
number of short phrases in these three languages.  We use the English
\textit{the president} and gender-specific expressions \textit{Mr President} and
\textit{Madam President} as well as gender-specific equivalents in French and
German.  The projection demonstrates a number of interesting results: First, the
model correctly clusters the words into three groups, corresponding to the three
English forms and their associated translations.  Second, a separation between
genders can be observed, with male forms on the bottom half of the chart and
female forms on the top, with the neutral \textit{the president} in the vertical
middle.  Finally, if we assume a horizontal line going through \textit{the
  president}, this line could be interpreted as a ``gender divide'', with male
and female versions of one expression mirroring each other on that line.  In the
case of \textit{the president} and its translations, this effect becomes even
clearer, with the neutral English expression being projected close to the
mid-point between each other language's gender-specific versions.

These results further support our hypothesis that the bilingual contrastive
error function can learn semantically plausible embeddings and furthermore, that
it can abstract away from mono-lingual surface realisations into a shared
semantic space across languages.

\section{Related Work}

\noindent\textbf{Distributed Representations}\hspace{0.1in}
Distributed representations can be learned through a number of approaches.  In
their simplest form, distributional information from large corpora can be used
to learn embeddings, where the words appearing within a certain window of the
target word are used to compute that word's embedding.  This is related to
topic-modelling techniques such as LSA \cite{Dumais:1988}, LSI, and LDA
\cite{Blei:2003}, but these methods use a document-level context, and tend to
capture the topics a word is used in rather than its more immediate syntactic
context.

Neural language models are another popular approach for inducing distributed
word representations \cite{Bengio:2003}.  They have received a lot of attention
in recent years \cite[\textit{inter
    alia}]{Collobert:2008,Mnih:2009,Mikolov:2010} and have achieved state of the
art performance in language modelling.
\newcite{Collobert:2011} further popularised using neural network architectures
for learning word embeddings from large amounts of largely unlabelled data by
showing the embeddings can then be used to improve standard supervised tasks.

Unsupervised word representations can easily be plugged into a variety of NLP
related tasks.  Tasks, where the use of distributed representations has resulted
in improvements include topic modelling \cite{Blei:2003} or named entity
recognition \cite{Turian:2010,Collobert:2011}.

\vspace{0.05in}
\noindent\textbf{Compositional Vector Models}\hspace{0.1in}
For a number of important problems, semantic representations of individual words
do not suffice, but instead a semantic representation of a larger
structure---e.g. a phrase or a sentence---is required.  Self-evidently, sparsity
prevents the learning of such representations using the same collocational
methods as applied to the word level.  Most literature instead focuses on
learning composition functions that represent the semantics of a larger
structure as a function of the representations of its parts.

Very simple composition functions have been shown to suffice for tasks such as
judging bigram semantic similarity \cite{Mitchell:2008}.  More complex
composition functions using matrix-vector composition, convolutional neural
networks or tensor composition have proved useful in tasks such as sentiment
analysis \cite{Socher:2011,Hermann:2013:ACL}, relational similarity
\cite{Turney:2012} or dialogue analysis \cite{Kalchbrenner:2013}.

\vspace{0.05in}
\noindent\textbf{Multilingual Representation Learning}\hspace{0.1in}
Most research on distributed representation induction has focused on single
languages.  English, with its large number of annotated resources, has enjoyed
most attention.  However, there exists a corpus of prior work on learning
multilingual embeddings or on using parallel data to transfer linguistic
information across languages.  One has to differentiate between approaches such
as \newcite{Al-Rfou:2013}, that learn embeddings across a large variety of
languages and models such as ours, that learn joint embeddings, that is a
projection into a shared semantic space across multiple languages.

Related to our work,
\newcite{Yih:2011} proposed S2Nets to learn joint embeddings of tf-idf vectors
for comparable documents. Their architecture optimises the cosine similarity of
documents, using relative semantic similarity scores during learning.
More recently, \newcite{Lauly:2013} proposed a bag-of-words
autoencoder model, where the bag-of-words representation in one language is used
to train the embeddings in another. By placing their vocabulary in a binary
branching tree, the probabilistic setup of this model is similar to that of
\newcite{Mnih:2009}. Similarly, \newcite{Chandar:2013} train a cross-lingual
encoder, where an autoencoder is used to recreate words in two languages in
parallel. This is effectively the linguistic extension of \newcite{Ngiam:2011},
who used a similar method for audio and video data.
\newcite{Hermann:2014:ICLR} propose a large-margin learner for multilingual word
representations, similar to the basic additive model proposed here, which, like
the approaches above, relies on a bag-of-words model for sentence
representations.

\newcite{Klementiev:2012}, our baseline in \S\ref{sec:rcv-cldc}, use a form of
multi-agent learning on word-aligned parallel data to transfer embeddings from
one language to another.
Earlier work, \newcite{Haghighi:2008}, proposed a method for inducing bilingual
lexica using monolingual feature representations and a small initial lexicon to
bootstrap with.  This approach has recently been extended by
\newcite{Mikolov:2013}, \newcite{Mikolov:2013a}, who developed a method for learning
transformation matrices to convert semantic vectors of one language into those
of another.  Is was demonstrated that this approach can be applied to improve
tasks related to machine translation.  Their CBOW model is also worth noting for
its similarities to the \addMod composition function used here.  Using a
slightly different approach, \newcite{Zou:2013}, also learned bilingual embeddings
for machine translation.

\section{Conclusion}

To summarize, we have presented a novel method for learning multilingual word
embeddings using parallel data in conjunction with a multilingual objective
function for compositional vector models. This approach extends the
distributional hypothesis to multilingual joint-space representations.  Coupled
with very simple composition functions, vectors learned with this method
outperform the state of the art on the task of cross-lingual document
classification.  Further experiments and analysis support our hypothesis that
bilingual signals are a useful tool for learning distributed representations by
enabling models to abstract away from mono-lingual surface realisations into a
deeper semantic space.

\section*{Acknowledgements}
This work was supported by a Xerox Foundation Award and EPSRC grant number
EP/K036580/1.

\bibliographystyle{acl}
\bibliography{../a.bib,../kmh_short.bib}

\end{document}